\pdfoutput=1
\documentclass[11pt]{scaleai-paper}

\usepackage{amsmath}
\usepackage{amsfonts}
\usepackage{amssymb}
\usepackage{amsthm}
\usepackage{booktabs}
\usepackage{tabularx}
\usepackage{tabulary}
\usepackage{multirow}
\usepackage{subcaption}
\usepackage{float}
\usepackage{nicefrac}
\usepackage{algorithm}
\usepackage{algpseudocode}
\usepackage{placeins}

\usepackage[round,sort&compress]{natbib}
\let\cite\citep
\usepackage{xspace}
\usepackage{url}
\usepackage[colorlinks=true,linkcolor=scaleLink,citecolor=scaleLink,urlcolor=scaleLink]{hyperref}
\usepackage[capitalise,nameinlink]{cleveref}

\usepackage{caption}
\newcolumntype{Y}{>{\RaggedRight\arraybackslash}X}

\papertype{Rubric-Guided Self-Distillation}
\contact{}

\title{Rubric-Guided Self-Distillation: \\ Post-Training Without Rubric Verifiers}

\author[1]{MohammadHossein Rezaei}
\author[1]{Anas Mahmoud}
\author[1]{Zihao Wang}
\author[1]{Utkarsh Tyagi}
\author[1]{Advait Gosai}
\author[1]{Razvan-Gabriel Dumitru}
\author[1]{Aakash Sabharwal}
\author[1]{Bing Liu}
\author[1]{Yunzhong He}
\affil[1]{Scale AI}

\begin{document}

\maketitle

\begin{abstract}
  Rubrics have emerged as an alternative to RLVR in open-ended domains where a single ground-truth final answer is not available. 
  Existing rubric-based training methods rely on an LLM verifier that scores each rollout against rubrics. 
  This introduces substantial training-time overhead, exposes optimization to verifier-specific biases, and reduces rubric feedback to a sparse end-of-trajectory signal.
  We propose \emph{Rubric-Guided Self-Distillation} (RGSD), a verifier-free training method in which the base policy, conditioned on the rubric, serves as the teacher for the unconditioned student. 
  RGSD distills the rubric-conditioned teacher distribution into the student token-by-token, replacing sparse trajectory-level rewards with dense per-token learning signals and removing the LLM judge from the training loop entirely.
  Across Qwen-2.5 (3B, 7B) and Qwen3-Thinking (4B, 8B) models on medical and science domains, 
  RGSD achieves rubric satisfaction comparable to judge-based GRPO while using one on-policy rollout per prompt and no training-time verifier calls. 
  Ablations show that raw rubrics provide a stronger teacher enrichment signal than self-generated reference responses, 
  while a stronger GRPO judge can outperform RGSD in some settings,
  positioning RGSD as a complementary verifier-free alternative when verifier cost or reliability is the bottleneck.
\end{abstract}

\section{Introduction}
\label{sec:introduction}

\begin{figure}[t!]
\centering
\includegraphics[width=1\linewidth]{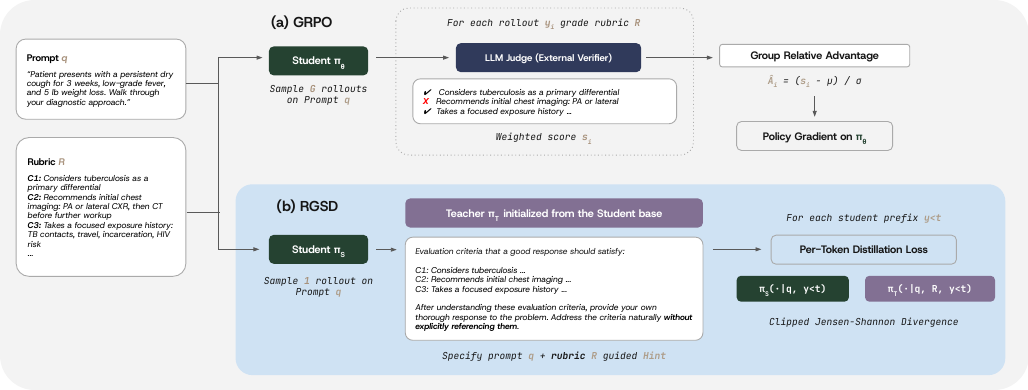}
\caption{\textbf{Method overview.} GRPO uses the rubric as an external grading signal: it samples \(G\) student rollouts per prompt, scores each rollout with an LLM judge, and converts the resulting scalar scores into group-relative policy-gradient updates. RGSD instead uses the rubric as privileged teacher context. The student samples one prompt-only rollout, while a frozen copy of the base model conditioned on the prompt and rubric provides next-token target distributions at each student-visited prefix. The student is trained with a clipped Jensen--Shannon distillation loss, removing the external judge from the training loop.}
\label{fig:method_overview}
\end{figure}

Reinforcement learning with verifiable rewards (RLVR) has driven much of the recent post-training progress on tasks where outcomes are automatically verifiable~\cite{shao2024deepseekmathpushinglimitsmathematical,deepseekr1}. 
However, for open-ended tasks such as clinical advice, scientific exposition, instruction following, and creative writing, programmatic verifiers are not applicable. 
To extend reinforcement learning to these domains, 
the community has converged on \emph{rubrics}: 
checklists of weighted criteria for each prompt.
Every response is graded against the rubric by an LLM verifier to produce a scalar reward~\cite{gunjal2025rubrics,viswanathan2025checklists,rezaei2025online,shao2025drtulureinforcementlearning}. 
Rubric-based RL has produced strong results and is increasingly becoming the default recipe for post-training on open-ended tasks.

Using LLMs as rubric verifiers is a powerful technique, but it introduces three costs:
(1) \emph{computational cost}: every rollout incurs a grading call to an LLM verifier, which is typically a computationally expensive LLM.
In addition to the grading cost itself, it makes the training loop inefficient as grading latency, rate limits, and failed calls stall the gradient update.
(2) \emph{verifier blind spots}: any reward signal derived from an LLM inherits the grader's idiosyncrasies and biases, so reward hacking can persist even with capable verifiers~\cite{mahmoud2026reward}.
(3) \emph{credit assignment}: rubric-based RL reduces a multi-criterion assessment of a response to a single scalar reward. While this reward is propagated to all generated tokens, it provides only a weak learning signal about which specific decisions improved or harmed rubric satisfaction, making credit assignment across long responses difficult.

We propose \textbf{Rubric-Guided Self-Distillation} (RGSD), 
a verifier-free training method for open-ended tasks evaluated through rubrics.
RGSD is motivated by the empirical observation that when a base model is shown the per-prompt rubric in its context, 
its responses score $30$--$45$pp higher than the unconditioned base (Table~\ref{tab:conditioning_gap}). 
RGSD turns this conditioning lift into a training signal through self-distillation.

At each training step, we sample an on-policy rollout from the \emph{student} (base model with only the prompt) 
and forward it through the \emph{teacher} (base model with the rubric placed in its context) to obtain a per-token target distribution.
We then update the student to match this target distribution under a clipped Jensen--Shannon divergence (Fig~\ref{fig:method_overview}).
RGSD removes the LLM judge from the training loop entirely and allows the model to internalize the rubrics through teacher conditioning rather than relying on external grading.

We evaluate RGSD across Qwen-2.5-Instruct (3B, 7B) and Qwen3-Thinking (4B, 8B), on two open-ended domains: medical and science. 
Results on in-distribution and out-of-distribution benchmarks show that RGSD is competitive with judge-based GRPO (on average $+6.1$ vs. $+5.9$pp on medical and $+4.9$ vs. $+4.5$pp on science),
while being more computationally efficient as it requires no judge calls and generates a single response instead of multiple rollouts.
Interestingly, we also observe that on Qwen-2.5-7B-Instruct, RGSD reaches peak quality at roughly half the response length of GRPO.

Finally, we ablate the GRPO judge strength and find that a stronger judge can outperform RGSD in some settings, at the cost of retaining the per-rollout verifier loop.
Also, we experiment with using a reference response in the teacher's context and find that rubrics provide a stronger enrichment.

Our contributions can be summarized as follows: 
(1) we introduce RGSD, a verifier-free training method for rubric-graded tasks, 
(2) we show that RGSD is competitive with judge-based GRPO across $4$ base models and $2$ domains, with no training-time verifier calls and substantially shorter responses on Qwen-2.5, and
(3) we ablate the enrichment signal and the GRPO judge strength, 
finding that the raw rubric is the stronger enrichment and that GRPO only outperforms RGSD when a more expensive judge is used.

\begin{table}[t]
\centering
\footnotesize
\setlength{\tabcolsep}{6pt}
\renewcommand{\arraystretch}{1.12}
\begin{tabular}{l cc cc}
\toprule
 & \multicolumn{2}{c}{\textit{RubricHub-medical}} & \multicolumn{2}{c}{\textit{RubricHub-science}} \\
\cmidrule(lr){2-3}\cmidrule(lr){4-5}
Base model & Base & + rubric in prompt & Base & + rubric in prompt \\
\midrule
Qwen-2.5-3B       & 27.0 & 71.0 \,(+44.0) & 31.0 & 62.0 \,(+31.0) \\
Qwen-2.5-7B       & 33.0 & 78.0 \,(+45.0) & 38.0 & 73.0 \,(+35.0) \\
Qwen3-4B-Thinking & 43.6 & 89.8 \,(+46.2) & 47.1 & 82.5 \,(+35.4) \\
Qwen3-8B-Thinking & 48.4 & 91.6 \,(+43.1) & 51.5 & 83.6 \,(+32.1) \\
\bottomrule
\end{tabular}
\caption{\textbf{Rubric-conditioning gap.} Rubric satisfaction score when the base model is asked to answer the prompt only (\emph{Base}) versus when the same base model is given the prompt together with the prompt-specific rubric set (\emph{+rubric in prompt}). 
The +rubric lift measures the rubric-conditioning gap: the amount by which the same base model improves when given the rubric as privileged context. We use this gap as a diagnostic for how much signal RGSD can potentially transfer into the unconditioned student.}
\label{tab:conditioning_gap}
\end{table}
\section{Related Work} 
\label{sec:related_work}

\paragraph{Rubrics for evaluating and training LLMs on open-ended tasks.}
Structured rubrics have become a common evaluation primitive for tasks where there is no programmatic verifier.
 Rubric-graded benchmarks now span clinical question answering~\cite{healthbench2025}, professional reasoning across law, finance, and other expert domains~\cite{prbench2025,profbench2025}, 
 instruction following~\cite{advancedif,multichallenge2025,audiomultichallenge2025}, long-form writing~\cite{writingbench2025}, 
 scholarly question answering~\cite{yifei2025researchqaevaluatingscholarlyquestion}, 
 and agentic tasks ranging from deep research~\cite{researchrubrics2025}
 to tool-use~\cite{bandi2026mcpatlaslargescalebenchmarktooluse}
 and software engineering~\cite{raghavendra2026sweatlasbenchmarkingcoding}.

 A parallel line of work uses these rubrics directly as the reward signal for reinforcement-learning post-training. 
 \emph{Rubrics as Rewards}~\cite{gunjal2025rubrics} aggregates multi-criterion rubric judgments into scalar rewards for GRPO and reports improvements over Likert-based reward baselines. 
 \citet{viswanathan2025checklists} extracts instruction-specific checklists per query and scores them with a mix of LLM judges and verifier programs. 
 Several efforts attack the rubric-construction problem itself by automating rubric generation at scale~\cite{li2026rubrichub,openrubrics,zhang2025chasing}, 
 or evolving the rubric set during training via pairwise comparison of rollouts~\cite{rezaei2025online}.
 \emph{POW3R}~\cite{tyagi2026rubricteachesequallypolicyaware} keeps the rubric fixed but re-weights them on-the-fly to upweight the criteria that better differentiate the rollouts.

A growing body of work characterizes how rubric-based RL can be exploited.
\citet{mahmoud2026reward} train policies against one rubric judge and evaluate them with a cross-family panel of frontier reference judges.
They observe that policies optimized against the training judge increasingly exploit judge-specific preferences, 
producing responses that receive high training-judge scores but are not preferred by the reference judge panel.
They also introduce a verifier-free \emph{self-internalization gap} diagnostic based on policy log-probabilities (similar to our teacher-student divergence),
which is highly correlated with the reference panel judge scores.
We exploit a closely related observation---the conditioning lift between an unconditioned and a rubric-conditioned base---but use it as a training signal rather than a diagnostic: 
RGSD distills the rubric-conditioned teacher into the unconditioned student.%

\paragraph{On-policy distillation for LLMs.} 
On-policy distillation traces back to imitation-learning work such as DAGGER~\cite{ross2011dagger},
which iteratively rolls out the student, and uses the expert's action at each visited state to update the student.
For language models, MiniLLM~\cite{gu2024minillm} re-popularized the idea with reverse-KL distillation between a small student and a large teacher. 
Generalized Knowledge Distillation~\cite{agarwal2024gkd} extends this to arbitrary divergences and on-policy student rollouts.

The most relevant works for RGSD are on-policy distillation efforts that quantify and refine the dense-versus-sparse advantage over RL. 
\citet{thinkingmachines2025opd} introduced On-Policy Distillation,
which distills a stronger teacher from the same model family into a student via a reverse-KL divergence under an on-policy student rollout.
They argue that an episode-level scalar reward provides $O(1)$ bits of supervision per episode, while a per-token distillation target provides $O(N)$. 
Therefore, on-policy distillation reaches the RL-trained policy in roughly $7$--$10\times$ fewer gradient steps and $50$--$100\times$ less cumulative compute.
On-Policy Distillation, however, still requires a stronger external teacher. 

\citet{zhao2026self} introduce \emph{On-Policy Self-Distillation} (OPSD),
in which the teacher and student are the same model under different conditioning contexts (the teacher is held frozen at the initial weights while the student is updated): 
the teacher conditions on privileged information such as a reference response or ground-truth answer, 
while the student sees only the question. 
Several concurrent works refine this teacher--student paradigm along different axes.
\citet{hubotter2026sdpo} condition the self-teacher on environment feedback (failing unit tests, runtime errors, judge critiques) 
and co-train teacher and student under EMA and trust-region regularization rather than freezing the teacher.
\citet{li2026rethinkingopd} complement these methods with a mechanistic study of on-policy distillation, 
showing it succeeds only when the teacher and student share compatible thinking patterns (i.e., high overlap ratio between student and teacher top-k tokens) and the teacher carries information beyond what the student has seen in training.

RGSD adopts On-Policy Distillation's dense per-token distillation and OPSD's self-distillation skeleton, 
but with a different teacher-conditioning signal: \emph{rubrics}.
Rubrics are available in open-ended domains where neither an external stronger teacher nor ground-truth reasoning traces exist.
To our knowledge, RGSD is the first to use rubrics as the teacher's enrichment signal in an on-policy self-distillation pipeline, 
and the first to position it as a promising alternative to verifier-based rewards in rubric-graded RL.

\section{Method}
\label{sec:method}

\subsection{Preliminaries}
\label{sec:method_prelim}

A training instance for rubric-graded open-ended generation is a tuple $(q, R)$, where $q$ is a prompt and $R = \{(c_i, w_i)\}_{i=1}^{K}$ is a rubric set: a list of criteria $c_i$ with weights $w_i$ describing what an ideal response should contain. Let $M_J$ denote an LLM judge. Given a candidate response $y$, the judge produces a binary satisfaction verdict $v_i(q, c_i, y) \in \{0, 1\}$ for each criterion $c_i$ (criterion $c_i$ is met or not). We aggregate these verdicts into the rubric score
\begin{equation}
    s_J(q, R, y) \;=\; \frac{\sum_{i=1}^{K} w_i \cdot v_i(q, c_i, y)}{\sum_{i=1}^{K} w_i} \;\in\; [0, 1].
    \label{eq:rubric_score}
\end{equation}
The per-criterion verdicts can be elicited either as $K$ separate judge calls (one criterion per call) or as a single batched call that returns the full $K$-element verdict for one rollout.
We use the batched variant throughout this paper (See Appendix~\ref{app:eval_prompt} for the judge prompt). The rubric-RL objective optimizes a policy $\pi_{\theta}$ to maximize expected rubric score:
\begin{equation}
    \max_{\theta} \;\; \mathbb{E}_{(q, R) \sim \mathcal{D}, \; y \sim \pi_{\theta}(\cdot \mid q)} \big[ s_J(q, R, y) \big].
    \label{eq:rubric_rl}
\end{equation}
GRPO~\cite{shao2024deepseekmathpushinglimitsmathematical} instantiates this by drawing $G$ rollouts per prompt, scoring each via $s_J$, and updating $\pi_{\theta}$ with a group-relative advantage estimate.
Every optimizer step thus requires $G$ batched judge calls per prompt (and $G \times K$ under per-criterion grading), which dominates training cost.

\subsection{Rubric-Guided Self-Distillation}
\label{sec:method_rgsd}

\begin{algorithm}[t]
    \caption{Rubric-Guided Self-Distillation (RGSD)}
    \label{alg:rgsd}
    \begin{algorithmic}[1]
        \Require Dataset $\mathcal{D} = \{(q_i, R_i)\}_{i=1}^{N}$; base model $\theta_{\text{base}}$; mixture coefficient $\beta$; JSD clip $\tau$; epochs $E$
        \State Initialize student $\theta_S \leftarrow \theta_{\text{base}}$; \;\; freeze teacher $\theta_T \leftarrow \theta_{\text{base}}$
        \For{epoch $= 1, \ldots, E$}
        \For{each minibatch $\mathcal{B} \subset \mathcal{D}$}
        \For{each $(q, R) \in \mathcal{B}$}
        \State Sample on-policy rollout $y \sim \pi_S(\cdot \mid q)$
        \State Compute teacher distribution $\pi_T(\cdot \mid q, R, y_{<t})$ for $t = 1, \ldots, T$
        \EndFor
        \State Update $\theta_S$ by descending $\nabla_{\theta_S}\, \mathcal{L}_{\text{RGSD}}$ \;\; (Eq.~\ref{eq:rgsd_loss})
        \EndFor
        \EndFor
    \end{algorithmic}
\end{algorithm}

RGSD (Figure~\ref{fig:method_overview}) removes the judge from the training loop.
Rather than \emph{grading} student rollouts with the judge model $M_J$ (Section~\ref{sec:method_prelim}),
we use the rubric to \emph{condition} the teacher whose distribution the student is trained to match.
RGSD instantiates two copies of the base model from the same checkpoint $\theta_{\text{base}}$:
a \emph{student} with trainable weights $\theta_S$ (initialized to $\theta_{\text{base}}$)
and a \emph{teacher} with frozen weights $\theta_T = \theta_{\text{base}}$.
The student conditions on only the prompt $q$ and defines a policy $\pi_S(\cdot \mid q)$ from which we sample on-policy rollouts during training.
The teacher conditions on the prompt $q$, the rubric set $R$, and the student's prefix $y_{<t}$ at each token position $t$, and defines a per-token distribution $\pi_T(\cdot \mid q, R, y_{<t})$ that we use as a distillation target.
The rubric is wrapped together with the prompt in a single user-turn message that ends with a short transition instructing the teacher to satisfy the rubrics without directly referencing them.
See Appendix~\ref{app:teacher_prompts} for the teacher input templates.

As shown in Algorithm~\ref{alg:rgsd}, a training step proceeds as follows:
We sample an on-policy rollout from the student,
$y \sim \pi_S(\cdot \mid q)$, then forward $(q, R, y)$ through the frozen teacher to obtain the per-token distribution $\pi_T(\cdot \mid q, R, y_{<t})$ at every position $t \in \{1, \ldots, T\}$.
The student is then updated to match the teacher token-by-token under a clipped Jensen--Shannon divergence:
\begin{equation}
    \mathcal{L}_{\text{RGSD}}(\theta_S) = \mathbb{E}_{(q, R) \sim \mathcal{D},\; y \sim \pi_S(\cdot\mid q)} \left[ \frac{1}{T} \sum_{t=1}^{T} D_{\beta}^{\text{clip}}\Big( \pi_S(\cdot \mid q, y_{<t}) \,\Big\|\, \pi_T(\cdot \mid q, R, y_{<t}) \Big) \right],
    \label{eq:rgsd_loss}
\end{equation}
where $D_{\beta}^{\text{clip}}$ is a clipped Jensen--Shannon divergence interpolating between forward KL ($\beta = 0$) and reverse KL ($\beta = 1$); we use $\beta = 0.5$ throughout. 
Only $\theta_S$ receives gradients; $\theta_T$ is held fixed for the entire training run, and the only purpose of the teacher forward is to provide a per-token target distribution. 
Importantly, this supervision is dense (every token in the rollout contributes a divergence term) rather than sparse (a single end-of-trajectory scalar),
and does not require any LLM judge or reward models during training.
The conditioning gap in Table~\ref{tab:conditioning_gap} provides a useful diagnostic for how much signal is available to transfer; Section~\ref{sec:results} analyzes the recovered fraction empirically.

When the base policy is a reasoning model, 
the teacher's reasoning trace might subtly refer to specific rubric criteria even when the system prompt explicitly asks it not to. 
Distilling those positions into a student that never sees the rubric could introduce noise or bias.
In order to prevent such leakage, we mask the tokens between \texttt{<think>} and \texttt{</think>} out of the loss, so only final response tokens contribute to the loss. 
This is a practical implementation detail that we ablate in Appendix~\ref{app:thinking_mask} and observe a $2$--$4$pp improvement on Qwen3-Thinking after masking vs. when the loss is applied to the entire trace.

\section{Experimental Setup}
\label{sec:setup}

\begin{table}[!t]
\centering
\footnotesize
\setlength{\tabcolsep}{4pt}
\renewcommand{\arraystretch}{1.12}
\begin{tabular}{l cccc}
\toprule
 & \multicolumn{2}{c}{\textit{Medical}} & \multicolumn{2}{c}{\textit{Science}} \\
\cmidrule(lr){2-3}\cmidrule(lr){4-5}
 & RubricHub-med & HealthBench & RubricHub-sci & ResearchQA \\
\midrule
Qwen-2.5-3B-Instruct & 19.9 & 19.3 & 24.7 & 54.9 \\
\quad + GRPO         & \textbf{28.5 (+8.6)} & \textbf{24.0 (+4.7)} & \textbf{27.2 (+2.5)} & \textbf{66.2 (+11.3)} \\
\quad + RGSD         & 27.9 (+8.0) & 23.6 (+4.3) & 27.1 (+2.4) & 65.5 (+10.6) \\
\midrule
Qwen-2.5-7B-Instruct & 25.1 & 23.5 & 32.4 & 59.6 \\
\quad + GRPO         & 35.5 (+10.4) & \textbf{32.9 (+9.4)} & \textbf{36.3 (+3.9)} & \textbf{69.3 (+9.7)} \\
\quad + RGSD         & \textbf{37.0 (+11.9)} & 30.1 (+6.6) & 35.9 (+3.5) & 68.2 (+8.6) \\
\midrule
Qwen3-4B-Thinking    & 34.5 & 34.2 & 41.7 & 66.5 \\
\quad + GRPO         & 39.8 (+5.3) & \textbf{37.6 (+3.4)} & 45.1 (+3.4) & 68.0 (+1.5) \\
\quad + RGSD         & \textbf{42.0 (+7.5)} & 34.2 (+0.0) & \textbf{45.5 (+3.8)} & \textbf{69.1 (+2.6)} \\
\midrule
Qwen3-8B-Thinking    & 39.2 & 39.6 & 46.0 & 69.5 \\
\quad + GRPO         & 43.2 (+4.0) & 41.4 (+1.8) & 47.7 (+1.7) & 71.2 (+1.7) \\
\quad + RGSD         & \textbf{47.1 (+7.9)} & \textbf{42.2 (+2.6)} & \textbf{50.3 (+4.3)} & \textbf{72.7 (+3.2)} \\
\midrule\midrule
\textit{Domain avg.\ $\Delta$pp} & \multicolumn{2}{c}{\textit{Medical}} & \multicolumn{2}{c}{\textit{Science}} \\
\quad + GRPO         & \multicolumn{2}{c}{+5.9} & \multicolumn{2}{c}{+4.5} \\
\quad + RGSD         & \multicolumn{2}{c}{\textbf{+6.1}} & \multicolumn{2}{c}{\textbf{+4.9}} \\
\bottomrule
\end{tabular}
\caption{\textbf{Main results.} Scores are rubric satisfaction percentages at the checkpoint selected by best in-domain RubricHub score. 
Parentheses report the improvement over the unconditioned base in percentage points. 
The domain-average rows average $\Delta$pp across all four base models and both benchmarks within each domain. 
\textbf{Bold} marks the higher score between GRPO and RGSD for each model-benchmark pair.}
\label{tab:main_results}
\end{table}

We compare RGSD against judge-based GRPO on the medical and science domains of RubricHub~\cite{li2026rubrichub}.
Each instance consists of a free-form prompt $q$ and a per-prompt rubric set $R$ of weighted criteria (See Appendix~\ref{app:data_examples}).
We train Qwen-2.5-3B/7B-Instruct~\cite{qwen2024qwen25} and Qwen3-4B/8B-Thinking~\cite{qwen2025qwen3} on the full train splits ($12{,}519$ medical and $19{,}806$ science prompts)
and evaluate on $300$-prompt subsets of each domain's held-out RubricHub validation split, plus $300$-prompt subsets of HealthBench~\cite{healthbench2025} for medical and ResearchQA~\cite{yifei2025researchqaevaluatingscholarlyquestion} for science as out-of-distribution benchmarks.
At evaluation time, all responses are graded against rubrics by gpt-5.4 at temperature $1.0$ using the judge prompt in Appendix~\ref{app:eval_prompt}.
We evaluate every $50$ optimizer steps and report each run's best in-domain RubricHub checkpoint (with every other benchmark scored at that same checkpoint).
GRPO uses $G=16$ rollouts per prompt and gpt-4o-mini as the training judge; RGSD samples one on-policy rollout per prompt and makes no judge calls during training.
Otherwise the two methods are matched on data, base model, optimizer, and schedule (5 epochs, effective batch size $128$, learning rate $4.2 \times 10^{-6}$).
For full hyperparameters, see Appendix~\ref{app:implementation}.

\begin{figure}[!t]
    \centering
    \includegraphics[width=\linewidth]{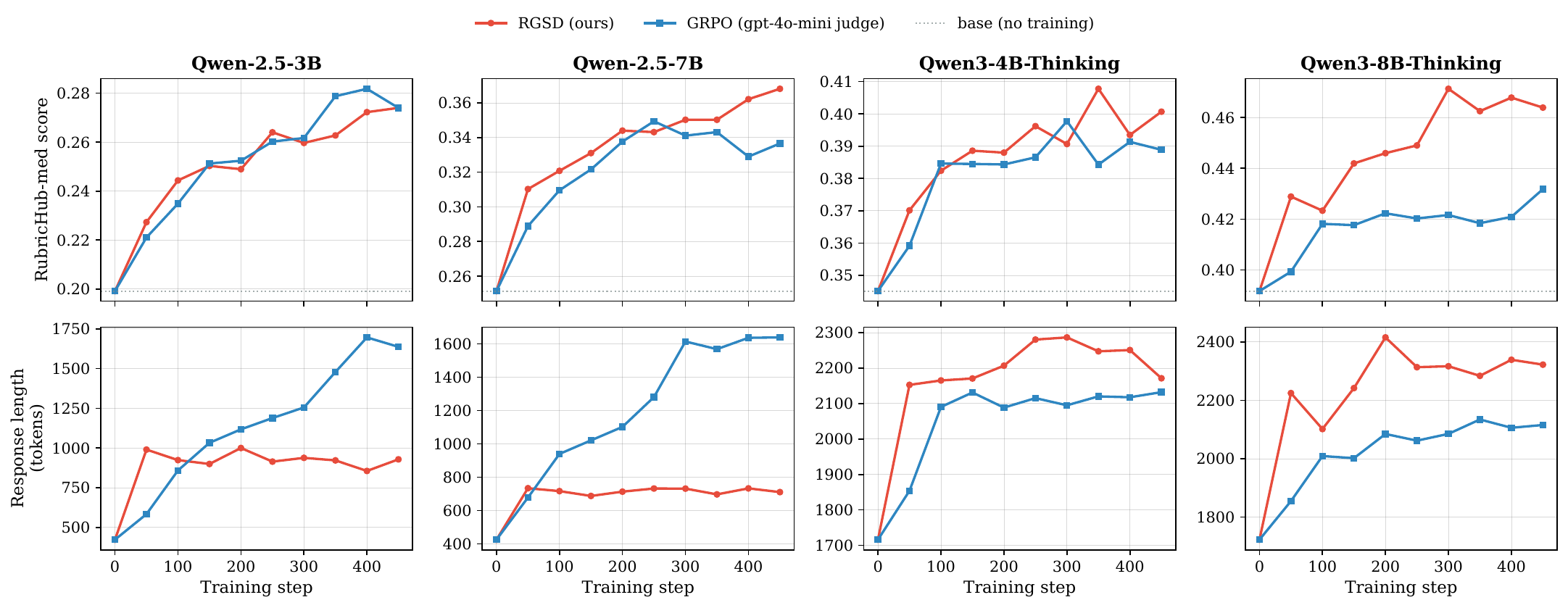}
    \caption{\textbf{Training dynamics on RubricHub-med-300.} Each column is a base model; the top row shows evaluation-time rubric satisfaction and the bottom row shows mean response length. 
On medical, RGSD reaches comparable or higher scores than GRPO while avoiding the severe Qwen-2.5 verbosity drift seen under judge-based training. 
For Qwen3-Thinking, both methods operate in a longer reasoning-trace regime, where RGSD often scores higher but is not uniformly shorter. 
Dotted lines mark the corresponding untrained base scores.}
    \label{fig:training_dynamics_medical}
\end{figure}

\section{Results and Discussion}
\label{sec:results}

Table~\ref{tab:main_results} compares RGSD with judge-based GRPO across four base models and two training domains.
RGSD achieves improvements comparable to GRPO in both domains.
Averaged over the two medical evaluations, RGSD improves over the base by +6.1 percentage points, compared with +5.9 for GRPO. 
On science, RGSD improves by +4.9 percentage points, compared with +4.5 for GRPO. 
The per-model differences are, however, mixed: 
RGSD obtains the largest positive margin on Qwen3-8B-Thinking in medical (+7.9pp vs. +4.0pp), 
whereas GRPO has its largest advantage on Qwen3-4B-Thinking on HealthBench (+3.4pp vs. +0.0pp).
We therefore interpret these results as evidence of quality parity rather than a consistent accuracy advantage for either method.

\paragraph{GRPO gains come at a cost.}The main distinction between the methods is training-time supervision cost. 
Under our five-epoch setup with 16 rollouts per prompt for GRPO, 
GRPO requires 80 batched judge calls per training prompt, 
corresponding to approximately 1.0M judge calls for the medical split (12,519 prompts) and 1.6M for the science split (19,806 prompts). 
RGSD makes no judge calls during training and samples one rollout per prompt per epoch. 
It instead obtains supervision from a frozen rubric-conditioned teacher distribution. 

We position RGSD as a verifier-free alternative to GRPO that matches its quality in this evaluation regime while eliminating training-time judge calls, 
not as a strict quality improvement over GRPO.
It is also worth noting that the results in Table~\ref{tab:main_results} are from single-seed runs.
A 3-seed envelope on our headline Qwen-2.5-7B-Instruct medical configuration in Appendix~\ref{app:multi_seed} shows tight seed variance ($\sigma\!\approx\!0.4$pp at peak).
Thus, we read sub-$1$pp differences as indicating parity rather than as significant wins.

\begin{figure}[!t]
\centering
\includegraphics[width=\linewidth]{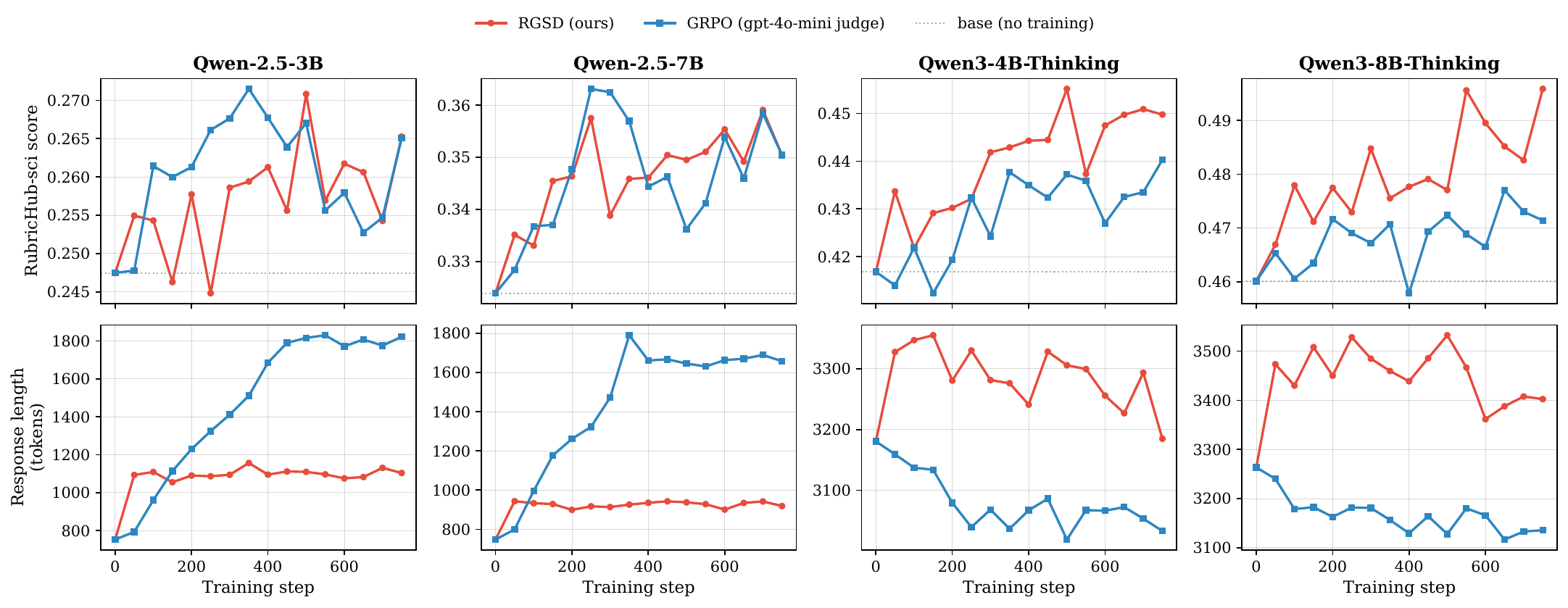}
\caption{\textbf{Training dynamics on RubricHub-sci-300.} Each column is a base model; the top row shows evaluation-time rubric satisfaction and the bottom row shows mean response length. 
On Qwen-2.5, GRPO again becomes much longer without a consistent score advantage over RGSD. 
On Qwen3-Thinking, RGSD has the stronger score trajectory, but length behavior differs from Qwen-2.5: GRPO shortens relative to the base while RGSD remains longer and scores higher. 
Dotted lines mark the corresponding untrained base scores.}
\label{fig:training_dynamics_science}
\end{figure}

\paragraph{RGSD is concise.}Figures~\ref{fig:training_dynamics_medical} and~\ref{fig:training_dynamics_science} show evaluation-time rubric satisfaction and mean response length over training. 
The strongest length effect appears on the Qwen-2.5 models, where RGSD reaches comparable peak rubric satisfaction with substantially shorter responses: 
at the best-scoring checkpoints, GRPO is $\sim\!1.4\text{--}1.8\times$ longer than RGSD, 
and by the final evaluated checkpoint it is $\sim\!1.6\text{--}2.3\times$ longer. 
This length growth does not yield a consistent score advantage, 
suggesting that judge-based optimization can reward coverage or thoroughness even when the extra tokens do not improve evaluation score.

The Qwen3-Thinking models behave differently. 
Their base responses are already longer because they include reasoning traces, and post-training length changes are smaller in relative terms. 
On medical, both methods lengthen by a few hundred tokens; on science, RGSD lengthens modestly while GRPO becomes shorter than the base. 
We therefore view the length result as family-dependent: 
RGSD avoids severe GRPO verbosity drift on Qwen-2.5, but it is not uniformly shorter on Thinking models.

\paragraph{GRPO leads to significantly more false claims.}Length inflation in rubric-based RL can be a symptom of reward hacking.
\citet{mahmoud2026reward} find
that optimizing presence-heavy rubric criteria is associated with longer, more claim-dense responses, 
with gains in completeness co-occurring with declines in factual correctness and conciseness under rubric-free evaluation.
We therefore ran an auxiliary HealthBench-300 factual-claim audit on the Qwen-2.5-7B medical run (Appendix~\ref{app:claim_audit}). 
For each checkpoint response, an LLM first extracted distinct verifiable medical claims 
and then verified each claim as \emph{correct}, \emph{incorrect}, or \emph{fabricated}. 
By the final audited checkpoint, GRPO increased from 15.8 to 38.4 claims per response and from 30.5\% to 45.1\% false-claim rate, 
whereas RGSD increased more modestly from 15.8 to 25.8 claims and from 30.5\% to 35.1\% false-claim rate. 
Since the audit is itself LLM-based, we treat it as supporting evidence rather than as a primary factuality benchmark.
With that caveat, the extra GRPO tokens appear to introduce additional unsupported claims, not merely more complete explanations.

The score trajectories provide a complementary view of optimization. 
In the Qwen-2.5 runs, GRPO reaches its best evaluation score before the end of training in all four trajectories and then partially regresses, 
most visibly on Qwen-2.5-7B: from peak to final, it drops by 1.3pp on medical and 2.7pp on science. 
RGSD is less affected in the same settings: on medical it continues improving through the final checkpoint for both Qwen-2.5 sizes, 
and on science it either peaks at the final checkpoint or remains within 0.8pp of its peak. 
For Qwen3-Thinking, the pattern is less a late-stage collapse than a persistent evaluation-score gap, 
with RGSD attaining the higher peak score on all four runs. 
These dynamics are consistent with a proxy-reward gap between the GRPO training judge (gpt-4o-mini) and the stronger evaluation judge (gpt-5.4) as characterized by \citet{mahmoud2026reward}: 
continued optimization against the training judge can favor response patterns that do not transfer to the evaluation judge. 
RGSD removes this particular scalar-reward optimization loop by matching a frozen rubric-conditioned teacher distribution, 
but its attainable improvement is bounded by the quality of that teacher.

The size of the rubric-conditioning gap in Table~\ref{tab:conditioning_gap} helps explain the domain-level differences. 
In medical, conditioning the base model on the rubric increases rubric satisfaction by roughly 43--46 percentage points, and RGSD recovers a substantial fraction of this gap (avg.\ $+8.8$pp on RubricHub-med, ${\sim}20\%$ of the conditioning lift).
In science, the conditioning gap is smaller, roughly 31--35 percentage points, and the observed RGSD gains are correspondingly smaller (avg.\ $+3.5$pp on RubricHub-sci, ${\sim}11\%$ of the lift). 
This suggests that the rubric-conditioning gap can serve as a lightweight, imperfect diagnostic for when RGSD is likely to be effective: 
if the base model benefits strongly from seeing the rubric at inference time, then self-distillation has more signal to transfer into the unconditioned student.

\subsection{Ablations}
\label{sec:ablations}

We ablate two design choices that determine when RGSD should be preferred over verifier-based training: the form of privileged context used by the teacher, and the quality of the GRPO training judge.

\begin{figure}[!t]
\centering
\includegraphics[width=\linewidth]{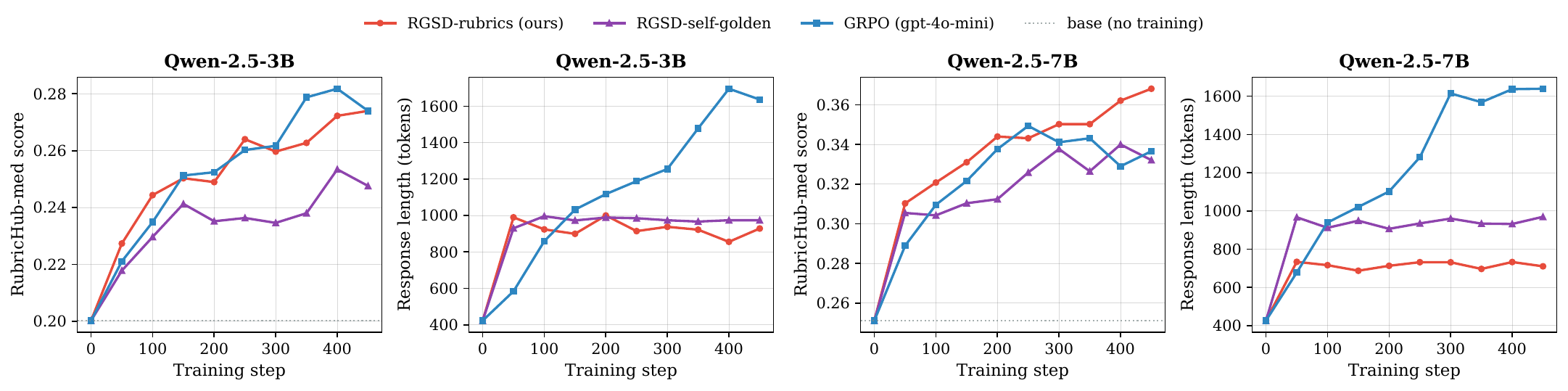}
\caption{\textbf{Enrichment ablation on RubricHub-med-300.} For each Qwen-2.5 size, we plot rubric satisfaction and mean response length for RGSD with raw rubrics, RGSD with a self-generated rubric-conditioned reference response, and GRPO with the default gpt-4o-mini judge. Raw rubrics provide the stronger enrichment signal. Both teacher-conditioning variants avoid the severe late-stage length growth of GRPO, though the self-golden variant is longer than raw-rubric RGSD.}
\label{fig:enrichment_ablation}
\end{figure}

\paragraph{Is the rubric itself the right enrichment signal?}
RGSD uses the raw rubric as the teacher's privileged context, but a natural alternative is to follow the self-distillation setup of OPSD more closely and condition the teacher on a reference response instead. This alternative is also empirically plausible in our setting: Table~\ref{tab:conditioning_gap} shows that rubric-conditioned base responses score substantially above unconditioned base responses, so these responses are reasonable candidates for ``self-golden'' supervision. We therefore compare the default rubric-conditioned teacher against a self-golden variant in which the teacher receives a rubric-conditioned response generated by the same base model, rather than the rubric itself.

Figure~\ref{fig:enrichment_ablation} shows that the raw rubric is the stronger enrichment signal. On Qwen-2.5-3B/7B-Instruct medical, rubric-RGSD peaks at 27.4/36.8 rubric satisfaction, compared with 25.3/34.0 for the self-golden variant, a gap of 2--3 percentage points. Both teacher-conditioning variants avoid the severe verbosity drift of GRPO, but the self-golden response does not improve the length-quality tradeoff; on Qwen-2.5-7B, it is both lower-scoring and longer than the raw-rubric teacher. We interpret this as a coverage effect. A single reference response is one sampled realization of the rubric: it may omit criteria, overcommit to one explanation structure, or encode stylistic artifacts of the sampled answer. The rubric, by contrast, preserves the full set of criteria and 
lets the frozen teacher decide which tokens satisfy them at each student-visited prefix. 
The rubric variant is also simpler operationally, since it uses the supervision already present in each training instance and avoids the additional offline generation pass required to construct self-golden responses. We therefore use  rubrics as the enrichment signal in all experiments.

\begin{figure}[!t]
\centering
\includegraphics[width=\linewidth]{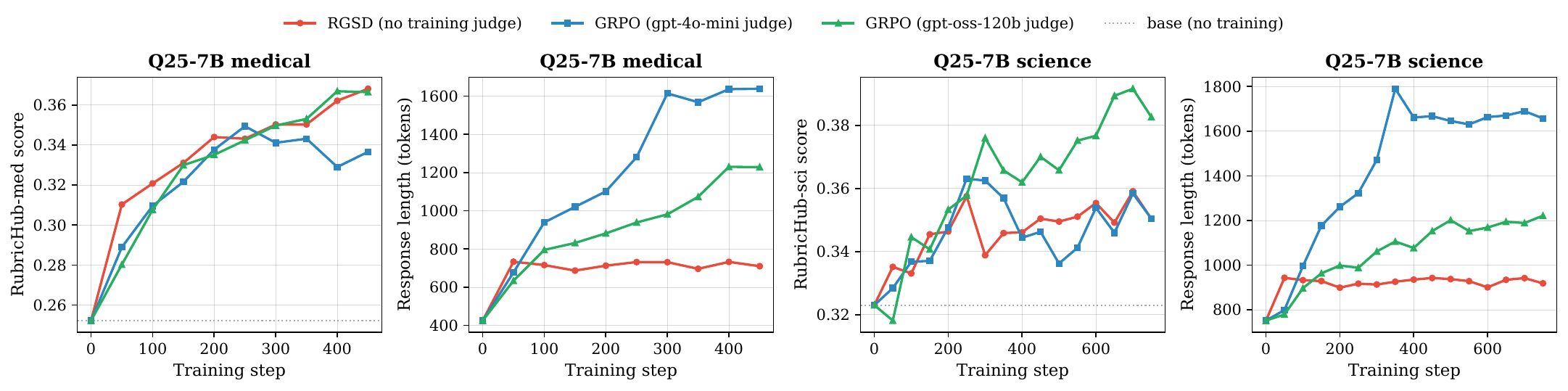}
\caption{\textbf{Judge-strength ablation.} For each domain, we plot primary RubricHub score and mean response length for RGSD with no training-time judge, GRPO with the default gpt-4o-mini judge, and GRPO with the stronger gpt-oss-120b judge. The stronger judge improves GRPO and surpasses RGSD on science, but both GRPO variants retain the per-rollout verifier loop and produce longer responses than RGSD.}
\label{fig:judge_ablation}
\end{figure}

\paragraph{Can GRPO close the gap with a stronger judge?}
Our main GRPO baseline uses gpt-4o-mini as the training judge, so Figure~\ref{fig:judge_ablation} asks whether the comparison changes when GRPO is trained with a stronger verifier. Replacing gpt-4o-mini with gpt-oss-120b improves GRPO substantially, but the effect is domain-dependent. On Qwen-2.5-7B medical, the stronger judge nearly matches RGSD on RubricHub-med, reaching 36.7 versus 36.8 at peak. On science, however, the stronger judge outperforms RGSD, reaching 39.2 versus 35.9 at peak on RubricHub-sci. This is an important caveat: RGSD should not be read as dominating the best attainable judge-based GRPO configuration. When a stronger verifier is available and the domain benefits from its additional signal, GRPO can achieve higher rubric satisfaction.

The tradeoff is cost and operational complexity. The stronger-judge GRPO run still requires verifier calls for every rollout during training, while every verifier call being more computationally expensive. 
It also remains longer than RGSD at inference: at the final checkpoint, GRPO with gpt-oss-120b judge produces roughly 1,200 tokens versus 700 for RGSD on medical, and 1,200 versus 900 on science. 
Consistent with our factual-claim audit,
\citet{mahmoud2026reward} found that even while using a stronger verifier, longer responses result in a higher false-claim rate.
Thus, the stronger judge changes the quality frontier but not the basic cost structure. We view RGSD and judge-based GRPO as complementary regimes: RGSD is attractive when verifier calls are expensive, unavailable, or unreliable, and when the base model exhibits a strong rubric-conditioning response; stronger-judge GRPO may be preferable when maximizing rubric score justifies the additional verifier cost.

\section{Conclusion}
\label{sec:conclusion}

We introduced Rubric-Guided Self-Distillation (RGSD), 
a verifier-free post-training method for rubric-graded open-ended generation. 
Instead of scoring each rollout with an external LLM judge, 
RGSD uses the base model conditioned on the prompt-specific rubric as a frozen teacher and trains the unconditioned student to match the teacher distribution token-by-token. 
This converts rubric supervision from a sparse trajectory-level scalar reward into a dense per-token distillation signal with no expensive verifier calls.

Across four base models and two rubric-graded domains, 
RGSD achieves comparable performance to judge-based GRPO, 
while requiring no training-time judge calls and only one on-policy rollout per prompt rather than sixteen. 
The comparison is not a uniform quality win for either method: 
average gains are close, per-model margins are mixed, and a stronger GRPO judge can outperform RGSD in some settings. 
The main advantage of RGSD is therefore not that it dominates verifier-based RL, 
but that it recovers much of its benefit without putting an LLM judge in the optimization loop.

The training dynamics suggest additional practical differences. 
On Qwen-2.5 models, RGSD avoids the severe verbosity drift observed under GRPO, 
reaching comparable rubric satisfaction with substantially shorter responses. 
Our auxiliary HealthBench factual-claim audit further suggests that GRPO's additional tokens can introduce more unsupported claims.
On Qwen3-Thinking models, length behavior is more domain-dependent because the base responses already contain long reasoning traces.

Overall, our results position RGSD and judge-based GRPO as complementary regimes. 
RGSD is attractive when verifier calls are expensive, unreliable, or operationally difficult, 
and when the base model exhibits a large rubric-conditioning response. 
Stronger-judge GRPO may be preferable when maximizing rubric score justifies the additional verifier cost. 
More broadly, RGSD shows that rubrics need not be used only as external grading instruments: 
they can also serve as privileged context for dense self-distillation.

\phantomsection
\addcontentsline{toc}{section}{Limitations}
\section*{Limitations}
\label{sec:limitations}

Our experiments are designed to compare RGSD against a verifier-based GRPO pipeline at realistic training scale, 
which makes exhaustive replication expensive. 
A five-epoch GRPO run with 16 rollouts requires 80 judge calls per training prompt, 
corresponding to roughly 1.0M judge calls on medical and 1.6M on science in our setup. 
For this reason, most main-table runs are single-seed. 
We partially mitigate this by evaluating a broad grid of four base models, two domains, and both in-domain and out-of-domain benchmarks, 
and by running a three-seed envelope on the headline Qwen-2.5-7B medical configuration. 
We therefore interpret small margins as parity rather than as statistically significant wins.

RGSD is most useful when the base model has a meaningful rubric-conditioning response. 
This is not an incidental weakness but the operating regime of the method: 
RGSD transfers behavior that the same base model can express when given the rubric as privileged context.
 The rubric-conditioning gap provides a cheap diagnostic before training, 
 since it can be measured with a forward-only validation pass rather than a full GRPO run. 
 When this gap is small, or when a substantially stronger verifier is available and affordable, 
 judge-based GRPO can be the better choice.
 Thus, RGSD should be viewed as complementary to verifier-based RL rather than as a universal replacement.

Our evaluation still uses LLM judges. 
RGSD removes the LLM verifier from the training loop, 
but rubric satisfaction at evaluation time is still measured by a stronger held-out judge, 
and the HealthBench factual-claim audit uses an LLM extract-and-verify pipeline. 
This matches current practice for rubric-graded open-ended evaluation 
and lets us test against a judge different from the GRPO training judge, 
but the absolute scores may still inherit judge-specific biases. 
We therefore treat the factual-claim audit as supporting evidence for the response-length analysis rather than as a standalone factuality benchmark.

Finally, reasoning-trace models require a small implementation adjustment. 
For Qwen3-Thinking, the rubric-conditioned teacher can refer to rubrics inside the thinking trace, 
while the student does not receive the rubric at inference time. 
We mask thinking-block tokens and distill only the final answer tokens, 
which improves performance in Appendix~\ref{app:thinking_mask}. 
This masking is a practical recipe choice for reasoning-trace bases, not a conceptual requirement of RGSD for standard instruction-tuned models.

\bibliographystyle{plainnat}

\appendix

\section{Implementation Details}
\label{app:implementation}

This appendix collects the configuration details required to reproduce the main-text results.

\paragraph{Hyperparameters.}
Table~\ref{tab:hyperparams} reports the full per-method configuration. Effective batch size, learning rate, schedule, and total optimizer step count are matched across RGSD and GRPO to support an apples-to-apples comparison; method-specific entries are marked ``---'' where not applicable.

\begin{table}[t]
\centering
\begin{tabular}{lcc}
\toprule
& RGSD & GRPO \\
\midrule
Parameter regime          & full-FT     & full-FT \\
Effective batch size      & 128         & 128 \\
Learning rate             & 4.2e-6      & 4.2e-6 \\
Schedule                  & const+warmup & const+warmup \\
Warmup ratio              & 0.10        & 0.10 \\
Epochs                    & 5           & 5 \\
Optimizer steps (med/sci) & 490 / 775   & 490 / 775 \\
Max gradient norm         & 0.10        & 0.10 \\
Gradient sharding         & ZeRO-3      & ZeRO-3 \\
Max sequence length       & 8192        & 8192 \\
Max completion length     & 2048 / 4096$^{\dagger}$ & 2048 / 4096$^{\dagger}$ \\
Distillation $\beta$      & 0.5         &--- \\
JSD clip $\tau$           & 0.05        &--- \\
Top-$k$ vocab for loss$^{\ddagger}$ & 128 &--- \\
Strip thinking-block loss & on (Qwen3-Thinking) &--- \\
Rollouts per prompt $G$   &---          & 16 \\
KL coefficient            &---          & 0.01 \\
Training judge            &---          & gpt-4o-mini (default; gpt-oss-120b in ablation) \\
\bottomrule
\end{tabular}
\\[0.3em]
{\footnotesize $^{\dagger}$ $2048$ tokens for instruction-tuned bases; $4096$ for Qwen3-Thinking bases.\\
$^{\ddagger}$ Restricting the JSD loss to the top-$128$ teacher vocabulary entries is required for Qwen3-8B-Thinking (vocab $\approx 152$K) to fit in $8\!\times\!80$GB H100s under ZeRO-3.}
\caption{Full hyperparameters across methods. Effective batch size, learning rate, and optimizer step count are identical across methods.}
\label{tab:hyperparams}
\end{table}

\paragraph{Hardware and software.}
Each training run uses a single $8\!\times\!$NVIDIA H100 node (80GB per GPU, NVLink-connected). The RGSD trainer is implemented on top of HuggingFace \texttt{transformers} with DeepSpeed ZeRO-3 for gradient sharding and vLLM-backed on-policy rollouts (vLLM $\geq 0.10$). The GRPO baseline uses the same trainer base; the training judge is served via an OpenAI-compatible LiteLLM gateway and called per rollout. The evaluation judge (gpt-5.4) is queried through the same gateway.

\paragraph{Training data.}
We train on the full train splits of RubricHub~\cite{li2026rubrichub} for each domain: $12{,}519$ prompts on medical and $19{,}806$ prompts on science. For evaluation we use disjoint $300$-prompt subsets of each domain's held-out validation split. Each instance carries the prompt $q$ and a rubric set $R$ (criterion text and weight). The self-golden enrichment ablation (Section~\ref{sec:results}) additionally consumes rubric-conditioned responses generated by the base model itself; we generate one self-golden per prompt under temperature $1.0$ and top-$p$ $0.95$ to encourage rubric coverage. Generation prompts and rubric-conditioning templates are reproduced verbatim in Appendix~\ref{app:prompts}. No prompt-deduplication or rubric filtering is applied beyond the RubricHub release defaults.

\paragraph{Conditioning-gap measurement protocol.}
The rubric-conditioning gap reported in Table~\ref{tab:conditioning_gap} is computed under a different protocol than the main results in Table~\ref{tab:main_results}, so the two tables' absolute base scores are not directly comparable.
Specifically, Table~\ref{tab:main_results} is scored by \texttt{gpt-5.4} on the $300$-prompt held-out validation subset of each domain, whereas Table~\ref{tab:conditioning_gap} is scored by \texttt{gpt-5.4-mini} on a $4{,}000$-prompt subset of the \emph{training} split.
The training-split scoring is a byproduct of generating the rubric-conditioned base responses used as self-golden teachers in the OPSD-style enrichment ablation (Section~\ref{sec:results}): we reuse those same responses both as the self-golden teacher inputs and as the ``+rubric in prompt'' column of Table~\ref{tab:conditioning_gap}, and grade them together with the unconditioned base responses on the same prompts.
We used the smaller \texttt{gpt-5.4-mini} judge for this pass because the volume is roughly $32$K per-prompt judgments per domain (4K prompts $\times$ 2 conditioning variants $\times$ 4 base models), which is an order of magnitude larger than the main evaluation grid.
\texttt{gpt-5.4-mini} grades more leniently than \texttt{gpt-5.4} across all base models, which accounts for the systematic offset in absolute scores between the two tables.
The relevant quantity in Table~\ref{tab:conditioning_gap} is the within-row \emph{lift} between the unconditioned base and the rubric-conditioned base (both scored by the same judge on the same prompts), which is what Section~\ref{sec:results} uses as a diagnostic for the RGSD-recoverable headroom.
\section{Additional Results}
\label{app:extra_results}

This appendix provides (a) the thinking-token mask ablation that justifies the recipe modification for reasoning-trace bases, with a case study of rubric leakage in the teacher's thinking trace; (b) the multi-seed envelope on the headline (base, domain) configuration; (c) a cross-family generalization study; and (d) a factual-claim audit of RGSD and GRPO responses on HealthBench-300.

\subsection{Thinking-Token Mask Ablation}
\label{app:thinking_mask}

\begin{figure}[t]
\centering
\includegraphics[width=\linewidth]{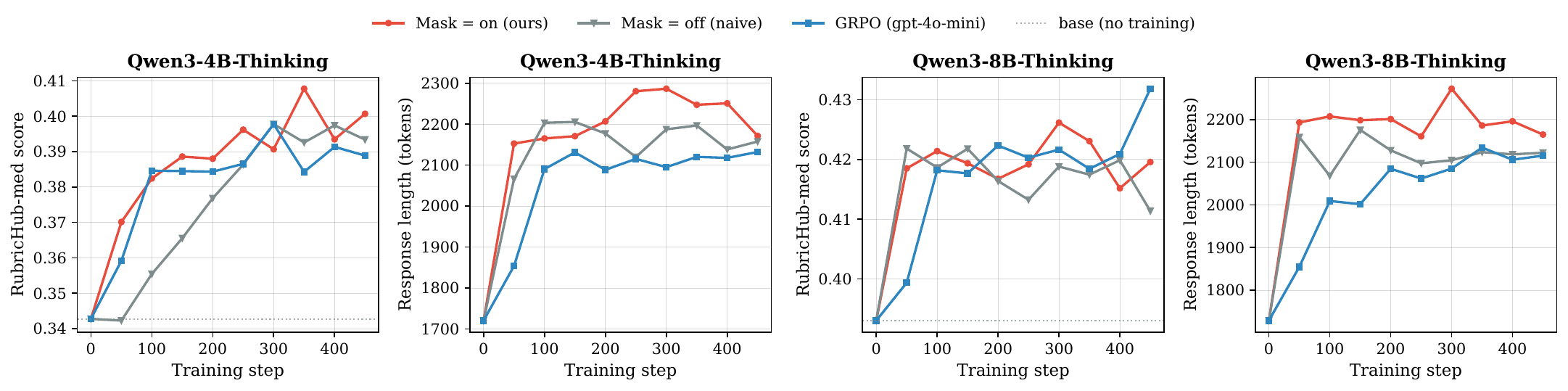}
\caption{\textbf{Thinking-token mask ablation on Qwen3-Thinking medical.} For each model size, the RubricHub-med score and the mean response length, under three regimes: mask\,=\,on (ours, red), mask\,=\,off (naive, gray), and GRPO+gpt-4o-mini (blue) for context. Within each model the ablation is clean (same enrichment); see caveat above for the across-size enrichment difference.}
\label{fig:thinking_mask}
\end{figure}

When applying RGSD to Qwen3-Thinking bases we adopt a two-part recipe: (i) mask all tokens between \texttt{<think>} and \texttt{</think>} out of the distillation loss, and (ii) extend the max-completion budget to $4$k tokens. Figure~\ref{fig:thinking_mask} ablates the masking component on both Qwen3-Thinking sizes (4B and 8B). Without the mask, the per-token loss includes the thinking trace---roughly $70\%$ of the generated tokens---which pulls the student toward matching the teacher's thinking style rather than its answer distribution. Across both sizes, mask\,=\,on outperforms mask\,=\,off on the primary score. We also overlay GRPO+gpt-4o-mini for context.

\paragraph{Why mask: a case study of rubric leakage.}
The teacher template (Figure~\ref{fig:teacher_template}) ends with an explicit instruction to ``\textit{address the criteria naturally without explicitly referencing them}''---yet the model's thinking trace routinely walks through the rubric criterion-by-criterion, often by literal number. Figure~\ref{fig:rubric_leakage_case} shows one representative Qwen3-4B-Thinking trace generated under the exact teacher template used during RGSD training. The leakage pattern is the rule, not the exception: scanning all $4{,}000$ rubric-conditioned medical-train teacher generations for an explicit rubric reference in the \texttt{<think>} block (regex over phrases like ``criterion $n$'', ``rubric'', ``evaluation criteria'', ``checklist'') gives:

\begin{center}
\small
\begin{tabular}{lcc}
\toprule
& Qwen3-4B-Thinking & Qwen3-8B-Thinking \\
\midrule
Rubric-conditioned (medical) & $70.2\%$ ($2810/4000$) & $73.0\%$ ($2920/4000$) \\
Rubric-conditioned (science) & $41.1\%$ ($1644/4000$) & $44.5\%$ ($1778/4000$) \\
Unconditioned\ \,(medical, $n=300$) & $\phantom{0}0.7\%$ & $\phantom{0}0.3\%$ \\
Unconditioned\ \,(science,\ \,$n=300$) & $\phantom{0}0.0\%$ & $\phantom{0}0.0\%$ \\
\bottomrule
\end{tabular}
\end{center}

\noindent The asymmetry is severe: $70$--$73\%$ of rubric-conditioned medical traces leak vs.\ near-zero on the matched unconditioned controls, even though the same teacher template instruction is active in both. Distilling these traces token-by-token into a student that \emph{never} sees the rubric pushes the student to produce thinking content that references criteria it has no access to---which is what motivates the mask.

\begin{figure}[t]
\centering
\begin{tcolorbox}[
  enhanced,
  colback=scalePanel,
  colframe=scaleSlate,
  boxrule=0.5pt,
  arc=3pt,
  left=10pt, right=10pt, top=6pt, bottom=6pt,
  title={\textbf{Rubric leakage in Qwen3-4B-Thinking trace}\,\textendash\,RubricHub-medical (instance \#2203)\,\textendash\,$30$ criteria},
  fonttitle=\small\bfseries,
  coltitle=white,
]
\small
\textbf{Prompt.} \textit{What are some characteristics and symptoms of thyrotoxic periodic paralysis?}

\smallskip
\textbf{Rubric criteria the teacher saw} (the same numbering the model uses below; first three of $30$):
\begin{enumerate}[leftmargin=2.2em, itemsep=1pt, topsep=2pt]
\item The response defines thyrotoxic periodic paralysis as episodic muscle weakness caused by hypokalemia occurring in the context of thyrotoxicosis.
\item The response explicitly links the condition to hyperthyroidism/thyrotoxicosis, noting that thyroid overactivity may be mild or subclinical.
\item The response mentions that TPP most commonly affects younger adult males, with higher prevalence in Asian, Hispanic, or Polynesian populations.
\end{enumerate}

\smallskip
\textbf{Thinking trace} (\texttt{<think>}\,\ldots\,\texttt{</think>}, excerpted; rubric-leakage spans in \textcolor{scaleTerracotta}{coral}):
\begin{quote}\small\itshape
Okay, the user wants me to provide a thorough response about thyrotoxic periodic paralysis (TPP) \textcolor{scaleTerracotta}{based on specific evaluation criteria}. Let me start by recalling what I know about TPP. \ldots\ So first, I need to define it as episodic muscle weakness due to hypokalemia in the context of thyrotoxicosis. \textcolor{scaleTerracotta}{That covers criterion 1.}

Next, I have to connect it to hyperthyroidism, noting that the thyroid overactivity might be mild or subclinical. \textcolor{scaleTerracotta}{That's criterion 2.} I should mention that it's more common in younger males, especially in certain ethnic groups\ldots \textcolor{scaleTerracotta}{That's criterion 3.} The attacks are acute and transient, with normal strength in between. \textcolor{scaleTerracotta}{Criterion 4.}

[\ldots the trace continues mapping ``Criterion 5'' through ``Criterion 30'' to specific clinical points\ldots]

Now, I need to structure this information in a natural, coherent way. \ldots Let me make sure \textcolor{scaleTerracotta}{I don't miss any of the criteria}, especially the ones about the triad, the safety notes, and \textcolor{scaleTerracotta}{the exact phrases like ``rebound hyperkalemia''}.
\end{quote}
\end{tcolorbox}
\caption{\textbf{Rubric leakage in a Qwen3-4B-Thinking rubric-conditioned generation.}
Despite the teacher template's closing instruction to ``\textit{address the criteria naturally without explicitly referencing them}'' (Figure~\ref{fig:teacher_template}), the model's \texttt{<think>} block enumerates the rubric criterion-by-criterion (``\textit{That's criterion 2.}''$\ldots$``\textit{Criterion 30.}'') and even cites verbatim weight-$10$ rubric phrasing (``\textit{rebound hyperkalemia}'') as a structural target. Distilling these tokens into a student that never sees the rubric would teach the student to fabricate criterion references; we mask the \texttt{<think>} span out of the loss.}
\label{fig:rubric_leakage_case}
\end{figure}

\emph{Enrichment caveat:} the paired mask-on/off runs differ in enrichment across model sizes---Qwen3-4B uses rubric enrichment (the canonical RGSD setup), Qwen3-8B uses self-golden enrichment (we did not run a rubrics-nomask Q3-8B). Within each model the ablation is clean: same enrichment, mask varied.

\subsection{Multi-Seed Envelope on the Headline Cell}
\label{app:multi_seed}

\begin{figure}[t]
\centering
\includegraphics[width=\linewidth]{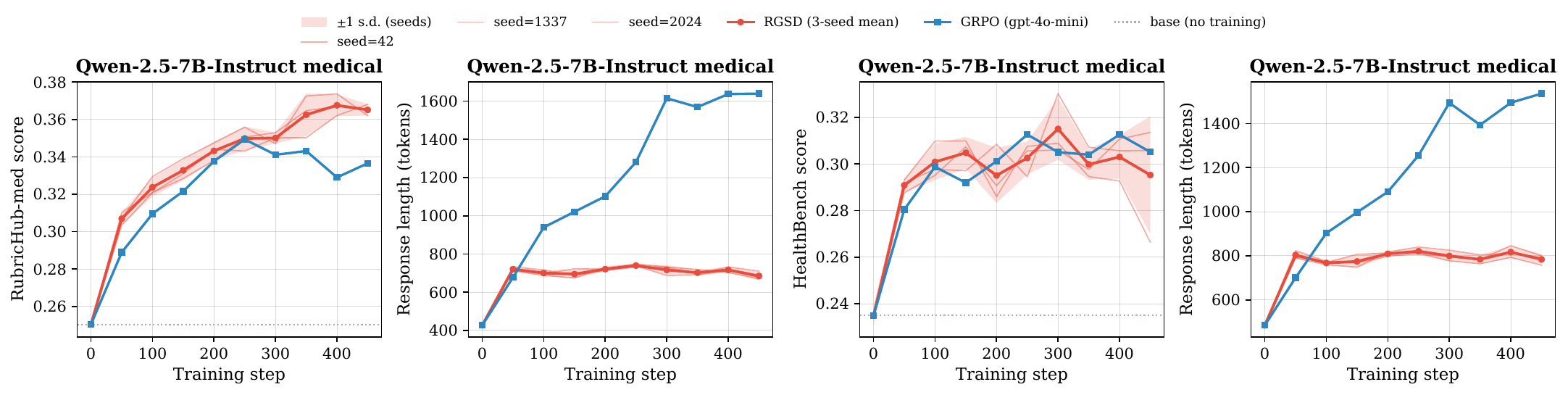}
\caption{\textbf{Multi-seed envelope on Qwen-2.5-7B-Instruct medical RGSD.} Light red lines: individual seed trajectories. Shaded red band: mean $\pm$ 1 s.d.\ across seeds. Bold red: 3-seed mean. Blue: GRPO+gpt-4o-mini for context. Dotted gray: base.}
\label{fig:multi_seed}
\end{figure}

Figure~\ref{fig:multi_seed} reports a 3-seed (42, 1337, 2024) envelope of RGSD on Qwen-2.5-7B-Instruct medical---the headline (base, domain) configuration of Table~\ref{tab:main_results}---to quantify training noise. The seed-mean $\pm$ 1 s.d.\ band on the primary RubricHub-med score is tight ($\sigma\!\approx\!0.4$pp at peak); the corresponding HealthBench band is wider but still well above the base. We plot single-seed runs for every other (base, domain) configuration in the main paper; this figure is the sanity check that the single-seed trajectories are not anomalous.

\subsection{Cross-Family Generalization}
\label{app:cross_family}

\begin{figure}[t]
\centering
\includegraphics[width=\linewidth]{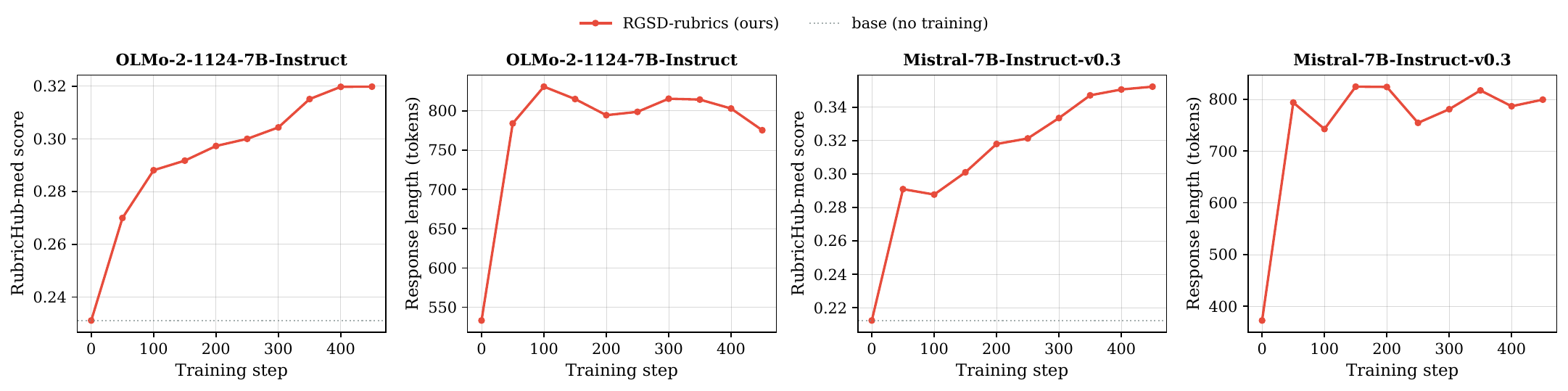}
\caption{\textbf{Cross-family RGSD training dynamics on RubricHub-med-300.} RubricHub-med score and mean response length over training for OLMo-2-1124-7B-Instruct and Mistral-7B-Instruct-v0.3 under RGSD. RGSD only (no GRPO baselines run for these families). Dotted gray marks the untrained base.}
\label{fig:cross_family_dynamics}
\end{figure}

The main paper focuses on the Qwen-2.5-Instruct and Qwen3-Thinking families. To check that RGSD is not specific to those families, Figure~\ref{fig:cross_family_dynamics} reports RGSD training dynamics on RubricHub-med for two additional bases under the same recipe: OLMo-2-1124-7B-Instruct and Mistral-7B-Instruct-v0.3. Both bases improve in rubric satisfaction over their unconditioned starting point over the course of training. We did not run matching GRPO baselines for these families because of the per-rollout judge cost; this study is only intended to confirm that the RGSD improvement transfers beyond the Qwen families, not to compare the two methods on them.

\subsection{Factual-Claim Audit on HealthBench-300}
\label{app:claim_audit}

Section~\ref{sec:results} observes that GRPO inflates response length on Qwen-2.5 medical without an accompanying gain on the rubric-graded score. We audit the \emph{content} of those longer responses on HealthBench-300 using an LLM extract-and-verify pipeline of the kind used by~\citet{mahmoud2026reward} to characterize rubric-RL reward hacking, run independently on our RGSD and GRPO checkpoints.

\paragraph{Pipeline.}
For each (prompt, response) pair, we prompt gpt-5.4 to extract every distinct verifiable claim from the response as a verbatim string---facts, statistics, mechanisms, drug indications, diagnostic criteria, named entities, and any other assertion that could be checked against medical knowledge. Generic advice (``see a doctor''), formatting markers, and exact-duplicate claims are not extracted. Each extracted claim is then verified in a second gpt-5.4 call against the model's parametric medical knowledge (no web search), assigning a verdict of \emph{correct} (factually accurate and precise), \emph{incorrect} (wrong, misleading, oversimplified, outdated, or missing critical caveats), or \emph{fabricated} (entirely invented; e.g., invented references or non-existent drugs). The per-checkpoint false-claim rate is the pooled $(\text{incorrect}+\text{fabricated})/\text{total claims}$ across the $300$ HealthBench prompts.

\paragraph{Results.}
Table~\ref{tab:claim_audit} reports per-checkpoint statistics for both methods on Qwen-2.5-7B-Instruct medical at training steps $0, 100, 200, 300, 400, 450$ (step $0$ is the unconditioned base; the small step-$0$ drift between the two columns is judge noise on the same set of base responses). GRPO's mean claim count per prompt grows steeply from $15.8$ (base) to $\sim\!38$--$40$ over training, and its false-claim rate climbs monotonically from $30.5\%$ to $45.1\%$. RGSD grows claim density more conservatively, saturating around $25$--$28$ claims per prompt, and its false-claim rate increases only modestly ($30.5\% \to 35.1\%$). At every checkpoint after step $0$, RGSD generates fewer claims, fewer false claims, and a lower false-claim percentage than GRPO.

\begin{table}[!t]
\centering
\footnotesize
\setlength{\tabcolsep}{4pt}
\renewcommand{\arraystretch}{1.05}
\begin{tabular}{r ccc ccc rrr}
\toprule
& \multicolumn{3}{c}{\textbf{RGSD}} & \multicolumn{3}{c}{\textbf{GRPO}} & \multicolumn{3}{c}{\textbf{RGSD $-$ GRPO}} \\
\cmidrule(lr){2-4}\cmidrule(lr){5-7}\cmidrule(lr){8-10}
Step & claims & false & false \% & claims & false & false \% & $\Delta$ claims & $\Delta$ false & $\Delta$ (pp) \\
\midrule
$0$   & $16.05$ & $4.91$  & $30.58\%$ & $15.84$ & $4.83$  & $30.51\%$ & $+0.20$   & $+0.07$ & $+0.07$  \\
$100$ & $25.64$ & $8.83$  & $34.46\%$ & $29.16$ & $11.68$ & $40.05\%$ & $-3.53$   & $-2.85$ & $-5.59$  \\
$200$ & $25.76$ & $9.66$  & $37.52\%$ & $33.49$ & $14.31$ & $42.71\%$ & $-7.74$   & $-4.64$ & $-5.20$  \\
$300$ & $24.78$ & $8.82$  & $35.61\%$ & $39.97$ & $17.50$ & $43.77\%$ & $-15.20$  & $-8.67$ & $-8.16$  \\
$400$ & $27.84$ & $10.50$ & $37.71\%$ & $39.60$ & $17.71$ & $44.72\%$ & $-11.75$  & $-7.21$ & $-7.01$  \\
$450$ & $25.80$ & $9.05$  & $35.08\%$ & $38.40$ & $17.33$ & $45.14\%$ & $-12.60$  & $-8.28$ & $-10.06$ \\
\bottomrule
\end{tabular}
\caption{\textbf{Factual-claim audit on HealthBench-300} for Qwen-2.5-7B-Instruct medical, by training step. ``claims'' is the mean total claims per prompt, ``false'' is the mean (incorrect $+$ fabricated) claims per prompt, ``false \%'' is the pooled false-claim rate across all $300$ prompts. Deltas are RGSD minus GRPO; the pp column is the difference in false \%.}
\label{tab:claim_audit}
\end{table}

\paragraph{Interpretation.}
GRPO's training judge (gpt-4o-mini) can credit a response for satisfying a rubric criterion even when the underlying claim is unsupported by the model's medical knowledge; over the course of training the policy discovers that producing more claims---true or false---raises the expected reward more reliably than carefully curating a smaller set of true claims. RGSD's distillation target is the rubric-conditioned base distribution rather than a scalar judge reward, so there is no gradient pushing the student toward additional unsupported content beyond what the rubric-conditioned base would have produced; the student inherits the base's claim density and accuracy profile. The audit complements the length-inflation observation in Section~\ref{sec:results}: the extra GRPO tokens are not just adding length but adding unsupported claims.
\section{Prompts and Data Examples}
\label{app:prompts}

This appendix reproduces the verbatim prompts used by the teacher and by the LLM judge, plus one sampled training instance per domain. We include them in full so that reviewers can sanity-check that no domain or task hint leaks through any of the conditioning channels.

\subsection{Teacher Input Template}
\label{app:teacher_prompts}
The student is always given only the user question $q$ wrapped in the model's standard chat template (no system prompt, no enrichment). The teacher receives $q$ wrapped together with the rubric as a single user-turn message; the verbatim template is reproduced in Figure~\ref{fig:teacher_template}. The closing instruction is what we call the \emph{transition prompt}: it tells the teacher to generate a fresh response \emph{conditioned on} the rubric rather than copying it. The transition is domain-agnostic; we use the same template for medical and science.

The self-golden ablation in Section~\ref{sec:results} replaces the rubric block with a pre-generated reference response from the same base model; that variant of the template is reproduced in Figure~\ref{fig:teacher_template_golden}.

\begin{figure}[t]
\centering
\begin{tcolorbox}[
  enhanced,
  colback=scalePanel,
  colframe=scaleSlate,
  boxrule=0.5pt,
  arc=3pt,
  left=10pt, right=10pt, top=6pt, bottom=6pt,
  title={\textbf{RGSD teacher template (rubric-conditioned)}},
  fonttitle=\small\bfseries,
  coltitle=white,
]
\small\itshape
\texttt{\{question\}}\\[0.5em]
Hidden evaluation criteria that a good response should satisfy:\\
1.~\texttt{\{criterion 1\}}\\
2.~\texttt{\{criterion 2\}}\\
\ldots\\
$K$.~\texttt{\{criterion K\}}\\[0.5em]
After understanding these evaluation criteria, provide your own thorough response to the problem. Address the criteria naturally without explicitly referencing them.
\end{tcolorbox}
\caption{\textbf{Teacher input template used at training time.} Given a question $q$ and a rubric set $R=\{(c_i,w_i)\}_{i=1}^K$, the teacher receives this single user-turn message. The closing instruction is the \emph{transition prompt}: it nudges the teacher to bake the rubric into the answer-shape rather than parroting it. The student receives only \texttt{\{question\}}.}
\label{fig:teacher_template}
\end{figure}

\begin{figure}[t]
\centering
\begin{tcolorbox}[
  enhanced,
  colback=scalePanel,
  colframe=scaleSlate,
  boxrule=0.5pt,
  arc=3pt,
  left=10pt, right=10pt, top=6pt, bottom=6pt,
  title={\textbf{Self-golden ablation teacher template} (used only in the self-golden enrichment ablation)},
  fonttitle=\small\bfseries,
  coltitle=white,
]
\small\itshape
\texttt{\{question\}}\\[0.5em]
Here is a reference response to this problem:\\
=== Reference Response Begin ===\\
\texttt{\{golden\_response\}}\\
=== Reference Response End ===\\[0.5em]
After understanding the reference response above, provide your own thorough response to the problem. Think step by step and be comprehensive.
\end{tcolorbox}
\caption{\textbf{Self-golden ablation teacher template.} For the enrichment-signal ablation in Section~\ref{sec:results}, the rubric block is replaced with a pre-generated rubric-conditioned reference response \texttt{\{golden\_response\}} from the same base model.}
\label{fig:teacher_template_golden}
\end{figure}

\subsection{Judge Prompt}
\label{app:eval_prompt}
At evaluation time we use gpt-5.4 (temperature $1.0$; the gpt-5 family requires temp $=1$) to score each response against the per-prompt rubric set; the same prompt structure is used for the GRPO training judge (gpt-4o-mini by default, gpt-oss-120b in the judge-strength ablation). The judge receives the prompt, the response, and the rubric criteria, and returns a binary satisfaction decision per criterion in a single batched call. The per-prompt reward is the weight-normalized fraction of satisfied criteria (Equation~\ref{eq:rubric_score}), identical in formula for training and evaluation. The verbatim prompt is reproduced in Figure~\ref{fig:judge_prompt}.

\begin{figure}[t]
\centering
\begin{tcolorbox}[
  enhanced,
  colback=scalePanel,
  colframe=scaleSlate,
  boxrule=0.5pt,
  arc=3pt,
  left=10pt, right=10pt, top=6pt, bottom=6pt,
  title={\textbf{Judge prompt template} (used at evaluation and for the GRPO training judge)},
  fonttitle=\small\bfseries,
  coltitle=white,
]
\small\itshape
You are an expert evaluator. Given a user question, a candidate response, and a list of evaluation criteria, decide for each criterion whether the response satisfies it.\\[0.4em]
=== Question ===\\
\texttt{\{question\}}\\[0.4em]
=== Response ===\\
\texttt{\{response\}}\\[0.4em]
=== Criteria ===\\
\texttt{\{numbered criterion list\}}\\[0.4em]
For each criterion, output a JSON object \texttt{\{"id": <int>, "satisfied": <true$|$false>, "reason": <one sentence>\}}. Output nothing other than a JSON array of these objects.
\end{tcolorbox}
\caption{\textbf{Judge prompt template.} The same prompt structure is used by the evaluation judge (gpt-5.4) and by the GRPO training judge (gpt-4o-mini by default; gpt-oss-120b in the judge-strength ablation in Section~\ref{sec:results}). The per-criterion verdicts are aggregated via Equation~\ref{eq:rubric_score} to produce the per-prompt reward.}
\label{fig:judge_prompt}
\end{figure}

\subsection{Sample Training Instances}
\label{app:data_examples}
Figures~\ref{fig:example_medical} and~\ref{fig:example_science} show one randomly-sampled training instance from each domain, with the question on top and the top-weighted criteria from the rubric set below. The student is given only the question; the teacher is given the question concatenated with the full rubric set, wrapped in the template of Figure~\ref{fig:teacher_template}.

\begin{figure}[t]
\centering
\begin{tcolorbox}[
  enhanced,
  colback=scalePanel,
  colframe=scaleSlate,
  boxrule=0.5pt,
  arc=3pt,
  left=10pt, right=10pt, top=6pt, bottom=6pt,
  title={\textbf{Medical}\,\textendash\,RubricHub-medical train (instance \#10476)\,\textendash\,\ensuremath{K\!=\!23} criteria, weights \ensuremath{\in [3,10]}},
  fonttitle=\small\bfseries,
  coltitle=white,
]
\small
\textbf{Prompt.} \textit{What disease is caused by a long-term lack of vitamin B1?}

\smallskip
\textbf{Rubric criteria} (six highest-weighted; weight column on the left):
\begin{itemize}[leftmargin=2.8em, labelsep=0.5em, itemsep=2pt, topsep=2pt, label={}]
\item[\textbf{w$\!=\!$10}] The response explicitly states that beriberi is the disease caused by long-term vitamin B1 (thiamine) deficiency.
\item[\textbf{w$\!=\!$10}] The response correctly states that the disease is beriberi and explicitly notes that vitamin B1 is also called thiamine.
\item[\textbf{w$\!=\!$\phantom{0}9}] The response explains both types of beriberi, describing wet beriberi as affecting the cardiovascular system and dry beriberi as affecting the nervous system.
\item[\textbf{w$\!=\!$\phantom{0}9}] The response includes a disclaimer advising the reader to seek immediate medical attention because the condition can be life-threatening.
\item[\textbf{w$\!=\!$\phantom{0}9}] The response mentions at least one risk factor for thiamine deficiency that is neither poor diet nor alcoholism (e.g., dialysis, malabsorption, bariatric surgery, chronic diarrhea).
\item[\textbf{w$\!=\!$\phantom{0}9}] For both wet and dry beriberi, the response lists at least four distinct symptoms, each presented as separate bullet points.
\end{itemize}
\end{tcolorbox}
\caption{\textbf{Sample medical training instance from RubricHub.} The student receives only the prompt; the judge scores each of the $K\!=\!23$ rubric criteria as met or not met, and the per-prompt reward is the weight-normalized fraction of satisfied criteria (Eq.~\ref{eq:rubric_score}). Six highest-weighted criteria are shown.}
\label{fig:example_medical}
\end{figure}

\begin{figure}[t]
\centering
\begin{tcolorbox}[
  enhanced,
  colback=scalePanel,
  colframe=scaleSlate,
  boxrule=0.5pt,
  arc=3pt,
  left=10pt, right=10pt, top=6pt, bottom=6pt,
  title={\textbf{Science}\,\textendash\,RubricHub-science train (instance \#7314)\,\textendash\,\ensuremath{K\!=\!22} criteria, weights \ensuremath{\in [3,12]}},
  fonttitle=\small\bfseries,
  coltitle=white,
]
\small
\textbf{Prompt.} \textit{Explain why intramolecular hydrogen bonding in molecules like ortho-nitro-phenol leads to a decrease in the melting point compared to molecules without such bonding, considering the effects on intermolecular interactions and molecular separation.}

\smallskip
\textbf{Rubric criteria} (six highest-weighted; weight column on the left):
\begin{itemize}[leftmargin=2.8em, labelsep=0.5em, itemsep=2pt, topsep=2pt, label={}]
\item[\textbf{w$\!=\!$12}] The response lists the experimentally measured melting point ($^\circ$C) for ortho-nitrophenol and for a comparable isomer that cannot form intramolecular hydrogen bonds (e.g., para-nitrophenol), with both values within $\pm 5\,^\circ$C of accepted literature.
\item[\textbf{w$\!=\!$11}] The response explicitly states that the intramolecular hydrogen bond forces the molecule into a bent or cyclic conformation that reduces crystal packing efficiency, and links this to the lower melting point.
\item[\textbf{w$\!=\!$10}] The response explicitly explains why intramolecular hydrogen bonding in ortho-nitrophenol lowers its melting point compared to molecules without such bonding.
\item[\textbf{w$\!=\!$10}] The response states that the intramolecular hydrogen bond consumes the hydrogen-bond donor/acceptor sites, reducing the molecule's ability to form intermolecular hydrogen bonds with neighboring molecules.
\item[\textbf{w$\!=\!$10}] The response links the weaker intermolecular forces to a lower amount of thermal energy required to disrupt the lattice, thereby explaining a lower melting point.
\item[\textbf{w$\!=\!$10}] The response describes the intramolecular hydrogen bond as forming a six-membered chelate ring and explicitly uses the term ``chelate ring'', correctly identifying the structural motif.
\end{itemize}
\end{tcolorbox}
\caption{\textbf{Sample science training instance from RubricHub.} Same convention as Figure~\ref{fig:example_medical}: the student sees only the prompt; the judge scores each of the $K\!=\!22$ criteria.}
\label{fig:example_science}
\end{figure}

\end{document}